# Influence of Swarm Intelligence in Data Clustering Mechanisms

Pitawelayalage Dasun Dileepa Pitawela[1], Gamage Upeksha Ganegoda[1]

[1]Faculty of Information Technology, University of Moratuwa, Sri Lanka
`dileepa.pitawela@gmail.com`, `upekshag@uom.lk`

**Abstract.** Data mining focuses on discovering interesting, non-trivial and meaningful information from large datasets. Data clustering is one of the unsupervised and descriptive data mining task which group data based on similarity features and physically stored together. As a partitioning clustering method, K-means is widely used due to its simplicity and easiness of implementation. But this method has limitations such as local optimal convergence and initial point sensibility. Due to these impediments, nature inspired Swarm based algorithms such as Artificial Bee Colony Algorithm, Ant Colony Optimization, Firefly Algorithm, Bat Algorithm and etc. are used for data clustering to cope with larger datasets with lack and inconsistency of data. In some cases, those algorithms are used with traditional approaches such as K-means as hybrid approaches to produce better results. This paper reviews the performances of these new approaches and compares which is best for certain problematic situation.

## 1 INTRODUCTION

Nature is always a source of inspiration. The techniques, structures, behaviors of nature are adapted in inventions, creations and in upgrading of present technology. Swarm Intelligence is one of the nature inspired technological aspect which influenced in data analysis.

When it comes to Information Technology era, data becomes the most powerful and rapidly generating unit of information. So it's a major aspect to analyze, classify and categorize data into separate groups in order to harness the efficiency of using most appropriate data in most applicable situation. One technique for this categorization is data clustering which is a task in data mining. Data clustering groups data, based on similarity features.

There are many mechanisms for cluster data and mainly they are divided into two categories as hierarchical and partitioning approaches which have been elaborated in literature review. Due to the limitations and drawbacks of used approaches, Swarm intelligence based mechanisms arises. This novel technique is more suitable due to its self-organized and decentralized behavior. Such behavior can be seen in ant colonies, in bee hives, in firefly behavior and etc. Ant Colony Optimization, Artificial Bee Colony Algorithm, Bat Algorithm, Firefly Algorithm are few example algorithms derives from swarm behavior.

This paper first describes those algorithms and then explains how they are used in data clustering. Then in discussion section, first each algorithm is quantitatively compared with conventional k-means clustering approach using benchmark datasets and then a qualitative comparison is given between Artificial Bee Colony algorithm, Firefly algorithm, Ant Colony Algorithm and Bat Algorithm.

## 2 LITERATURE REVIEW

In literature, it's been using many approaches to deal with data clustering and those methods are mainly divided into two categories as partitioning methods and hierarchical methods. Partitioning algorithms initially defines the number of clusters and then iteratively allocate data items among clusters. On the other hand, hierarchical algorithms divide or combine existing clusters and create a hierarchical structure which reveals the order in which groups are divided or merged. Based on literature reviews, the partitioning algorithms perform better than the hierarchical algorithms [1]. Most used data partitioning clustering algorithm is K-means clustering. Although it is easy to implemented and worked with, it has a major drawback. That is called as initial point sensibility. In other words, final solution is fully dependent on initially defined number of clusters and Euclidean distances associated among them.

### 2.1 K-means Clustering

This clustering mechanism follows a simple way to categorize a given dataset through a certain number of initially fixed clusters (assume k number of clusters). First it has to be defined a centroid for each cluster and then those centroids have to be placed in a greedy way since different locations causes different results at the end of the process. Therefore, the better way is to place them as much as possible far away from each other. Then each data point is taken from the dataset and associated with the nearest centroid. When all the data points are associated with a centroid, initial grouping is done. At this point, new locations for defined centroids are calculated as barycenters of clusters resulted from preceding step and for each data point, association is done again. This centroid location calculation and association is done until centroids stop their location change. Finally data points associated with clusters are identified and algorithm ends [2].

The below figure, fig. 1 shows the steps of K-means clustering. The stars represent centroids and circles represent data points in that figure. The third and fourth steps are looped until Centroid locations are fixed.

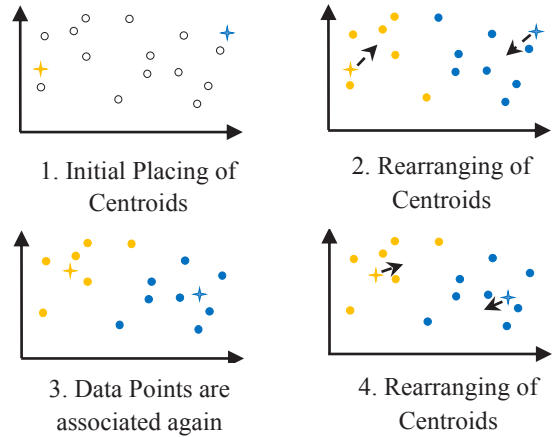

**Fig. 1.** Steps of K-mean Clustering

Although this is fast and efficient, the final result totally depends on the initial number of clusters and this mechanism performs low with big datasets. That's when the need of hybrid algorithms with swarm intelligence arise.

## 3  Swarm Intelligence Based Algorithms

The algorithms which have been developed based on Swarm Intelligence are known as swarm intelligence based algorithms. They are typically having inherent behavior of a Swarm. These algorithms work as agent based systems in order to depict the collective behavior of a swarm towards a single goal. Swarm Intelligence based algorithms perform better than normal algorithms in some contexts due to their self-organization and decentralized behavior.

This section discusses about swarm intelligence based algorithms. Since the context is data clustering, four major algorithms from swarm intelligence based algorithms which are being used for data clustering are taken to the consideration. They are, Firefly Algorithm, Ant Colony Algorithm, Artificial Bee Colony Algorithm and Bat Algorithm.

### 3.1 Firefly Algorithm

Since the results of K-means mechanism highly depend on the initial clusters, Firefly Algorithm (FA) has been proposed to identify initial clusters and then to use K-means to discover final solution or the final cluster set [3].

In Firefly algorithm, there are three idealized rules [4]:

1. Hence all fireflies are unisex, one will be attracted to another regardless of their sex.
2. Brightness of a firefly is directly proportional to their attractiveness. Therefore, for any two flashing fireflies, the less bright one will move towards the brighter one. The both attractiveness and brightness decrease as their distance increase. If there wasn't a brighter one compared to a particular firefly, it starts moving randomly.
3. The brightness of a firefly is determined by the landscape of the objective function. For a maximization problem, the brightness can simply be proportional to the value of the objective function.

In data clustering the objective function is a function which defines by the user in order to calculate the fitness value of a data point.

FA is an agent or population based optimization method inspired by the behavior of fireflies which move towards their flashing light. The characteristics of FA are [3],

- The location of each firefly corresponds to a problem state in the state space and its light intensity shows the fitness value of the position.
- The displacement and the direction of movement of a firefly is static to get attraction of neighboring fireflies.
- Fireflies move or fly in random directions if no attractive fireflies are around.

**Model of Firefly.**

First, each firefly, $s_i$ ($i = 1,…, N_f$) in FA has assignment $s_i[j]$ of cluster numbers which corresponds to data, which in turns relates to position vector of a firefly in the multi-dimensional space. The firefly also retains its fitness value or evaluation value $f(s_i)$ based on similarity between data. N denotes the number of fireflies and the data size or in other words, number of data items to be clustered [3].

**Movement of fireflies.**

Movement of a firefly can be depicted as an update or alteration of values of $s_i[j]$. Fireflies move towards the other which has more light intensity or in other words, more fitness value. The movement of a Firefly $i$, which is attracted to another more attractive Firefly $j$ is determined by the equation (1): [3]

$$X_i = x_i + \beta_0 \, e^{-\gamma r^2_{ij}} (x_i - x_j) + \alpha \left(\text{rand} - \frac{1}{2}\right) \quad (1)$$

While the second term corresponds to the attraction, the third term depicts randomization including randomization parameter $\alpha$ with the random number generator rand. The generated random numbers are uniformly distributed in the interval of [0, 1]. For the most cases of implementations, $\beta_0 = 1$ and $\alpha \in [0, 1]$. The parameter $\gamma$ represents the variation of the attractiveness. It varies between 0.01 and 100 in most scenarios [3]. Its value is important to determine the speed of the convergence and to understand how FA behaves [3].

The execution of firefly algorithm results a set of clustering points (centroids) which can be used as the input for the k-mean process. Subsequently, the k-mean process is carried on and this hybrid approach has been performed better than standalone k-mean process [5, 6]. It's been discovered that the proposed method is more effective for more complex datasets when number of attributes associated with data items [3]. Evaluation between both standalone and hybrid approach is on evaluation section.

### 3.2 Artificial Bee Colony Algorithm

This method is focuses on the accuracy of the clustering process of K-means. Identifying the correct cluster for a data item can be difficult due to the complexity of the dataset. Secondly, irrelevant cluster can be assigned because of having noise data to develop the objective function. Thirdly, generated rules for clustering do not represent the modeled system completely due to not having sufficient collection of data.

In Artificial Bee Colony (ABC) algorithm, three types of worker honeybees are responsible for food transportation to the hive [7, 8]. The first type is employed forager that has associated with a particular food source. They share information in the dancing area of the nest. The second type is scout bee that has been searching the environment surrounding the nest for finding new food sources. The third is onlooker bee waiting in the nest to learn food source position from any of employed forager which shares its own food source position and nectar amount by waggle dance. ABC algorithm contains five phases. They are initialization phase, employed bee phase, onlooker bee phase, memorize best solution phase and scout bee phase [7]. A graphical representation of ABC algorithm is in fig. 2.

In the *Initialization Phase*, each employed bee is assigned a random food source around the hive. That means number of employed bees in the hive is equal to the number of food sources around the hive.

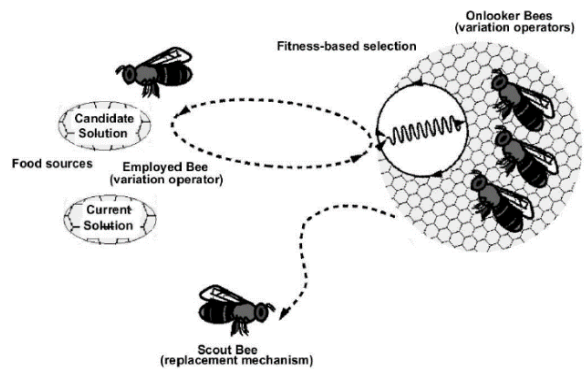

**Fig. 2.** Graphical representation of the elements in the ABC algorithm [9].

In the second, *Employed Bee Phase*, every employed bee searches for a better food source near the neighborhood area of its assigned food source. Employed bees memorize new food sources found in this run. Then every employed bee evaluates new food source against the previously assigned food source using the measure of nectar amount in each source. Then takes probabilistic decisions depending on the results of evaluation such that, if the new source has equal or more amount of nectar compared to the previous source, then new source is memorized over the old one in memory, or else, the old one is retained. A greedy selection mechanism is employed to obtain the decision of selecting old or new food source [7].

When it comes to the third, Onlooker Bee phase, after finishing search by employed bees, they share information in the dancing area of the hive. An onlooker bee identifies its previous food source by employed bees by making probabilistic decisions. Then onlooker bees make modifications to the food source by considering new source information brought by employed bees and check its fitness. After considering the nectar amounts of both sources, better source is memorized by an onlooker [7].

In the fourth, Memorize Best Source Phase, all the new food sources found in this iteration and in any previous iteration and their corresponding fitness values are compared and the best source is memorized. If the new best food source has better nectar amount than the previous best food source, new source is memorized. Otherwise, previous bee information is kept.

In the final, Scout Bee Phase, a new random food source is determined by Scout bees if nectar of a food source is consumed [7].

Every bee's food source corresponds to a possible centroid and nectar amount of the food source is corresponds to the fitness value of the source which is determined by the objective function. The clustering rule determination is described below in 4 steps as an analogy to basic ABC algorithm.

1. In the *Initialization Phase*, continuous and discrete values of the training set are normalized within a range [0, maxValue]. The obtained normalized values of training dataset are compared with the maxValue to determine whether each obtained value is in the rule or not. If an attribute's normalized value is larger compared to the maxValue, then it's not included in the proposed rule [7, 10].

2. In *Rule Discovery Phase,* a new rule is discovered. As the new rule's consequent part, the identified class which contains highest number of data from the training set is selected. After finding the new rule, correctly classified instances are removed from the training set [7]. There is a parameter max_unc_num (maximum uncovered examples ratio) which stops new rule discovering process. If the maximum uncovered instances are greater than max_unc_num, the algorithm stops creating a new classification rule. Finally unnecessary attributes are removed after the new rule is being analyzed [7].

3. The third, *Default rule phase*. In here, a default rule is determined for the remaining dataset. The default rule hasn't got an antecedent. As the consequent part of the default rule, the most frequent class in the remaining instances set is chosen [7].

4. In the Final, *Removing Unnecessary Rule Phase*, discovered rules are analyzed and the algorithm outputs the discovered rule set [7].

Based on these discovered rules, partitioning is done. Evaluation of basic k-mean clustering and hybrid approach with ABC algorithm is in the evaluation section.

### 3.3 Ant Colony Algorithm

Ant Colony Optimization algorithm was proposed by Marco Dorigo et al. It's based on natural way of ants determining shortest path from their colony to food source. Despite knowing individual ant has limited abilities, they can produce impressive group results such as nest building, defense, forming bridges, searching food items and etc. Searching food is the behavior which we are interested in this optimization.

Ant's communication is done using chemicals called pheromones. Social behavior of ants is driven by trail pheromones as they mark a path on ground using pheromones so that other ants can follow that path to find way to the food source. Normally, ants tend to follow trails with higher pheromone levels with a higher probability than trails with lower pheromone levels. Moreover when using the same path by more ants, as every ant releases pheromones, the pheromone density of a particular path gets high. This results in autocatalytic behavior, as more ants follow the trail, the trail becomes increasingly attractive.

In first, ant's food determining tactic is mentioned and then the applicability of ant colony optimization for data clustering is discussed.

Initially, with the aim of finding a food source, ants tend to move in all possible directions. If an ant found a food source F at some distance from its colony C, it releases pheromones while returning back to the colony. When the time passes, more ants use that pheromone path to reach the food source and they also release pheromones on their way. As a result, the intensity of the pheromone deposition gets stronger for the shorter path than the longer paths since ants always find the shorter path to the food source form the colony. Eventually, since most of the ants tend to follow the shortest path, the pheromone trails of longer paths get weak [11]. Ants shortest path convergence is shown in fig. 3. This unique behavior of ants is practiced to solve real world scenarios such as data clustering in data mining field.

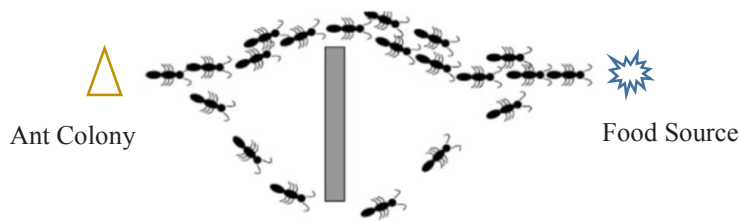

**Fig. 3.** Graphical representation of Ants converging to the shorter path rather than to longer path.

In order to apply Ant Colony Optimization (ACO) Algorithm to data clustering, it has been merged with K-means data clustering method. The hybrid algorithm of K-means combined ACO, results in better performance than when taken K-means and ACO algorithms individually. The problems such as mis-clustering of K-means and the problem of slow convergence of ACO has been addresses and overcome in the new hybrid algorithm [12].

ACO is used here as a refining algorithm for the results of K-means. So that, results of K-means is the input to the ACO. The main reason for this refinement is, by using any clustering algorithm, 100% accurate clustering result cannot be obtained [13]. There can be some errors known as mis-clustering. That means, data items which have been wrongly clustered. That type of errors can be reduced with the help of a refining algorithm.

The ants are allowed to walk randomly across the clusters. When an ant crosses a cluster, it picks an item from a cluster and drops it into another cluster. This pick and drop probability is calculated by equation (2) and (3) respectively [14].

$$\text{Picking up Probability, Pp} = \left(\frac{k_1}{k_1+f}\right)^2 \quad (2)$$

$$\text{Dropping Probability, Pd} = \left(\frac{f}{k_2+f}\right)^2 \quad (3)$$

Here, $k_1$ and $k_2$ are picking up threshold and dropping threshold respectively and f represents entropy value calculates before an item was picked and dropped [14]. If the picking probability is higher than the dropping probability, then the item is dropped in another cluster and the entropy value is calculated again for the cluster [14]. Entropy value of the cluster is somewhat like mean of the data values belong to the cluster.

ACO runs N times and N is specified by the controller of the algorithm according to their convenience. The results obtained using this algorithm are given in the evaluations.

### 3.4 Bat Algorithm

Bats are nocturnal mammals which can sense its prey in complete darkness. They have this ability since their body is capable of emitting natural electromagnetic rays and capturing the echo returning after encountering an obstacle or prey. Microbats, a species of Bats which produce a very loud sound which echoes back to them with exact same frequency [15]. Moreover, even if it's the same frequency, they have the ability to distinguish between their prey and other obstacles in the environment. Developing bat algorithm was inspired by this natural phenomenon of searching food by bats.

In this algorithm, each virtual bat flies on random direction in some velocity with a varying frequency and loudness towards a prey (a solution in data clustering). At the moment they find their prey, the frequency, loudness and pulse emission rate change. This process of searching the best solution continues until a certain stopping criteria met.

This traditional bat algorithm has been modified in order to apply for data clustering. Initially, each bat is assigned a cluster. Then the fitness value of the centroid in each bat is computed [1]. Next, based on the computed fitness value of centroid in a bat, the data items are divided and placed in proper cluster [1]. In successive iterations, the velocity, frequency and centroid values of bats are updated and a new solution is generated. Best solution is selected for each bat from the set of best solutions from the other bats. Increase of frequency and reduction of loudness are the criteria considered when accepting a new solution.

## 4     DISCUSSION

In this section, two major objectives are about to be covered. To remind, this paper is to determine that use of swarm intelligence based algorithms to data clustering is more applicable and efficient rather than using conventional k-means process. In order to prove that, in the first section of the discussion, performance of every swarm based algorithm is evaluated quantitatively and compared with the performance of k-means process separately. And then in the second section, with the intention of giving a general idea about the performances of swarm based approaches on one another, a limited qualitative evaluation is done based on few datasets commonly applied for each algorithm.

**4.1 Individual Critical Evaluations of Each Algorithm**

**Evaluation of Firefly Algorithm.**

In order to evaluate the proposed method, few experiments were done using benchmark datasets such as Iris dataset which contains details of three flowers (with 150 data and 4 attributes), Glass dataset which contains data about oxide content depending on the glass type (with 214 data and 9 attributes), Cancer-Int data set which contains data about diagnosis of Breast Cancer Wisconsin (original) (with 699 data and 9 attributes) and Haberman dataset which contains data on survival of patients who had undergone surgery for breast cancer (with 306 data and 3 attributes). For the parameters of the proposed method, the number of fireflies Nf was taken as 20 and the number of generations T was assigned as 100. The parameters α, β and γ were fixed to be 0.5, 1.0 and 1.0 respectively [3] (according to the equation (1)).

Table 1. Results on the K-means process [3]

|            | *IRIS* | *GLASS* | *CANCER-INT* | *HABERMAN* |
|---|---|---|---|---|
| *PCA*       | 85.1  | 21.7  | 51.0  | 50.7  |
| *TIME(SEC.)* | 0.069 | 0.024 | 0.058 | 0.023 |

Table 2. Results on Hybrid Method [3].

|  | IRIS | GLASS | CANCER-INT | HABERMAN |
|---|---|---|---|---|
| **PCA** | 88.3 | 38.9 | 96.0 | 58.3 |
| **TIME(SEC.)** | 0.22 | 0.38 | 0.64 | 0.26 |

In each of which result (In table 1 and table 2) depicts the average of the percentage of correct answers (PCA) and consumed time (in seconds) for 100 trials for each dataset. Despite of the requirement of slightly more time by the proposed method (according to table 2) compared to the k-means method (regarding table 1), the proposed method can find more suitable clusters since the percentage of correct answers is higher in proposed hybrid algorithm than k-means method after applying to every dataset.

**Evaluation of Artificial Bee Colony Method.**

To evaluate the performance of proposed Artificial Bee Colony algorithm, eleven benchmark datasets has been used. Those are Glass, Iris, E-coli, Wine, Liver disorder, Vowel, WDBC, Pima, CMC and two artificial datasets (Art1 and Art2) [16].

Table 3 shows the mean and the standard deviation obtained as a result of applying k-means process and ABC on previously mentioned benchmark datasets.

Table 3. Results of applying k-means clustering and ABC process separately for each dataset [16]

| *Dataset* | *K-Means* Mean Deviation | *ABC* Mean Deviation |
|---|---|---|
| *Art1* | 8.86E-01 0 | 8.86-01 5.71E-04 |
| *Art2* | 8.03E+00 1.34E+00 | 7.06E+00 3.56E-02 |
| *Iris* | 7.00E-01 8.03E-02 | 6.49E-01 1.30E-03 |
| *Wine* | 9.46E+01 1.31E+01 | 9.22E+01 4.56E-01 |
| *Glass* | 1.05E+00 6.03E-02 | 1.06E+00 1.83E-02 |
| *Ecoli* | 1.93E-01 3.31E-03 | 2.01E-01 1.79E-03 |
| *Liver Disorder* | 2.96E+01 0 | 2.86E+01 2.89E-03 |
| *Vowel* | 1.76E+02 <br> 4.02E+00 | 1.73E+02 <br> 1.02E+00 |
| *Pima* | 6.78E+01 <br> 1.46E-14 | 6.20E+01 <br> 2.74E-02 |
| *WDBC* | 2.68E+02 <br> 6.03E+01 | 2.67E+02 <br> 5.93E+01 |
| *CMC* | 3.76E+00 <br> 8.99E+01 | 3.81E+00 <br> 8.97E+01 |

Table 4 shows the given rankings for both algorithms considering the obtained results.

**Table 4.** Ranking of k-means and ABC process according to the results obtained from previous process [16]

| *Algorithm* | *Ranking* |
|---|---|
| K-Means | 5.05 |
| ABC | 3.09 |

Table 5 represents the computation time of each algorithm on each dataset. In other words, it shows the time taken for each algorithm to converge.

**Table 5.** Comparison of Execution time (sec.) on each benchmark dataset by both approaches [16]

| *Dataset* | *K-Means* | *ABC* |
|---|---|---|
| *Art1* | 0.92 | 0.83 |
| *Art2* | 0.63 | 0.57 |
| *Iris* | 0.35 | 0.27 |
| *Wine* | 1.06 | 0.79 |
| *Glass* | 1.53 | 1.25 |
| *Ecoli* | 2.48 | 2.04 |
| *Liver Disorder* | 0.73 | 0.61 |
| *Vowel* | 1.10 | 2.15 |
| *Pima* | 2.27 | 1.70 |
| *WDBC* | 5.58 | 3.94 |
| *CMC* | 6.93 | 4.85 |

By considering the standard deviations calculated from the results obtained by applying k-means and bee colony algorithms separately (as in table 3), it can be observed that almost always the standard deviation is lesser when the bee colony algorithm is used. Moreover considering the time taken to convergence (according to table 5), or in other words, considering the time taken to give the results in table 3, ABC algorithm takes lesser time compared to the k-means process. In addition to them, according to the ranks given through the results (in table 4), it can be concluded that ABC algorithm is more efficient than the k-means since ABC capable of providing better results in less amount of time [17, 18].

**Evaluation of Ant Colony Algorithm.**

In order to evaluate clustering results of Ant Colony algorithm, the results of K-means and Ant Colony Algorithm are compared here using two benchmark datasets. They are, Wisconsin Breast Cancer dataset (contains 569 instances with 32 attributes in each) and Dermatology dataset (contains 366 instances with 33 attributes in each).

Entropy value and the F-measure of the both obtained result sets are used for the comparison. [19].

- Entropy – At first, the class distribution of data is calculated for each cluster. Then the entropy for each cluster is calculated using the class distribution [19].
- F measure – Precision (also called positive predictive or in other words retrieved results that are relevant) and Recall (also called sensitivity in other words, fraction of relevant instances that are retrieved) values are sometimes used together to provide a single measurement to the system. It's called F-measure.

**Table 6.** Comparison of entropy values and values of F-Measure obtained from both k-means and ACO [19]

|  | K – Means | K – Means with ACO |
|---|---|---|
| *Wisconsin Breast Cancer Dataset* | | |
| No. of Classes | 2 | 2 |
| No. of Clusters | 2 | 2 |
| Entropy | 0.2373 | 0.1502 |
| F - measure | 0.9599 | 0.9799 |
| *Dermatology Dataset* | | |
| No. of Classes | 6 | 6 |
| No. of Clusters | 6 | 6 |
| Entropy | 0.0868 | 0.0103 |
| F - measure | 0.8303 | 0.8841 |
| *Thyroid Disease Dataset* | | |
| No. of Classes | 3 | 3 |
| No. of Clusters | 3 | 3 |
| Entropy | 0.0700 | 0.0530 |
| F – measure | 0.9103 | 0.9241 |

When entropy value gets higher, it implies that there is more unpredictability in the events being measured. So that it's better when entropy value gets lower [20, 21]. When the value of the F-Measure gets higher, it implies that the number of correct positive results obtained is higher. So that, it's better when the F-Measure (f-score) gets higher [22, 23].

By considering the results given in the table 6, it can be seen for the both datasets that the entropy value is lower in ACO compared to k – means and the f-measure is higher in ACO compared to k-means. So that it can be concluded that the use of ACO to refine results of k-means is better than using k-means as it is.

**Evaluation of Bat Algorithm.**

Evaluation of bat algorithm has been done based on the F-measure value calculated for several benchmark datasets by using k-means and bat algorithm. They are Iris (with 150 data and 4 attributes), Cancer (with 699 data and 9 attributes), Wine (consists 178 instances with 13 attributes in each) and Glass (with 214 data and 9 attributes). Results obtained are shown below in table 7.

In here, for each dataset, those algorithms are applied 10 times and average of calculated F-measure values is presented with the F-measure value of the best and the worst solution obtained within 10 runs [24].

Table 7. Results obtained by applying k-means and bat algorithm for clustering four benchmark datasets [24].

| *Dataset* | *Criteria* | *K-Means* | *Bat Algorithm* |
|---|---|---|---|
| *Iris* | Average | 0.868 | 0.890 |
|  | (Std) | (0.069) | (0.026) |
|  | Best | 0.892 | 0.926 |
|  | Worst | 0.670 | 0.832 |
| *Cancer* | Average | 0.961 | 0.958 |
|  | (Std) | (0.001) | (0.006) |
|  | Best | 0.962 | 0.966 |
|  | Worst | 0.960 | 0.950 |
| *Wine* | Average | 0.707 | 0.720 |
|  | (Std) | (0.025) | (0.008) |
|  | Best | 0.715 | 0.729 |
|  | Worst | 0.636 | 0.703 |
| *Glass* | Average | 0.519 | 0.505 |
|  | (Std) | (0.034) | (0.032) |
|  | Best | 0.557 | 0.551 |
|  | Worst | 0.456 | 0.466 |

F-measure values are taken here for the comparison, so that results are better if the F-measure value gets higher [25]. When observing table 7, it can be seen that for almost every instance, F-measure value obtained using bat algorithm is higher than obtained value using k-means.

The significance in the result set is that the F-measure value obtained is appeared to be higher even for the worst case solutions in almost every datasets. So that, it can be concluded that using bat algorithm provides better clustering results than using k-means.

### 4.2 A Qualitative Evaluation of Algorithms Taken Together

According to the usage, the applicability varies. In other words, despite the efficiencies of novel swam intelligence based algorithms than traditional approaches, performance of each algorithm depends on the situation it is used.

Firefly Algorithm commonly used to cluster large datasets since it performs well with particularly large datasets. By considering table 2 which shows percentage of correct answers for four datasets, the highest percentage is recorder for the cancer dataset which has highest number of records among the compared datasets there (dataset properties are mentioned above the table 1).

Researchers have found that the ABC algorithm performs better than other clustering algorithms in the situations which the dataset is not consistent, or in other words, when it's having noise data or lost values.

It's been discovered that ACO can be mainly used to cluster image type data. Normal numerical or string type datasets also use as inputs but for image clustering this algorithm performs better than others.

Although, so as to use bat algorithm, there weren't any specific reason, this algorithm is capable of performing clustering in low and high dimensional datasets such as Iris and Wine respectively as can be seen in table 7.

Table 8 represents a brief description about the most specific areas which the discussed algorithms can be applied effectively.

**Table 8.** Swarm intelligence based algorithm and it's most applicable area is shown respectively

| Swarm Intelligence based Algorithm | Prominent Applicable Area |
|---|---|
| Firefly Algorithm | Commonly used for large datasets which clustering points/centroids cannot be identified in earlier stages. |
| Artificial Bee Colony Algorithm | Applied in situations which the dataset is not consistent and it contains some lost values. |
| Ant Colony Optimization | Used as a refinement algorithm for k-means, when it's used to cluster normal datasets such as Iris. But ACO is specifically used in image data clustering. |
| Bat Algorithm | It hasn't got a specific situation to be used and it performs well with low dimensional datasets like Iris and high dimensional datasets such as Wine either. |

Based on the observed results in the individual comparison of each algorithm in the first section of the discussion, quantitative evaluation, it can be seen that the use of swarm intelligence based algorithms can lead the clustering process to a more effective and efficient way than using conventional k-means approach.

But an overall quantitative evaluation using all the algorithms cannot be done since different algorithms are evaluated based on different benchmark datasets. Moreover, efficiency and accuracy of different algorithms has been measured based on different criteria such as F-measure, entropy, standard deviation, mean, time taken to convergence and etc. So that, as mentioned above in the table 8, a qualitative approach has been taken.

This discussion was elaborated as two parts. The first contained individual quantitative evaluation of each selected algorithm as a comparison with k-means traditional clustering. But, because of the limitation mentioned in the above paragraph, an overall comparison taking all the algorithms together to a single dataset couldn't have been performed. The second part of the discussion continued as a qualitative evaluation of selected algorithms considering type of data which each algorithm can be applied on. So finally, based on the evaluation results, it can be concluded that the swarm based approaches performs better than traditional approaches in general and selecting the applicable algorithm depends on the type of the dataset.

## 5   CONCLUSION

In this paper, it has been discussed about traditional approaches of data clustering such as k-means and modern hybrid data clustering mechanisms such as Firefly Algorithm, Artificial Bee Colony Algorithm, Ant Colony Algorithm and Bat Algorithm emerged as alternatives for k-means. Moreover, it's proven that those novel approaches perform well compared to traditional approach in a quantitative manner. In addition to those, a qualitative comparison also has been given for discussed swarm intelligence algorithms.

By these evaluation and comparisons, it was expected to determine which approach, for instance, k-means or swarm based algorithms, are better for data clustering depending on the type of the data. By considering the approaches in the quantitative evaluation such as applying swarm based algorithms as it is or applying them together with k-means and considering the appropriate type of data for each algorithm as summarized in table 8 in qualitative evaluation, it can be concluded that the applicability of swarm based algorithms depends on the type of data and their effectiveness on a particular dataset can be vary according to the approach taken. So the whole purpose of this review, the identification of applicability of swarm based algorithms has been discovered.

As further enhancements, it's proposed to widen the scope by comparing those swarm based approaches together in a quantitative manner.

**Acknowledgement.** This review was supervised by Dr. Upeksha Ganegoda, senior lecturer, Department of Computational Mathematics, Faculty of Information Technology, University of Moratuwa. Special thanks to her for providing insights and sharing expertise towards the success of this review. Most information included here are from IEEE research papers.

## References

1. S. U. Mane and P. G. Gaikwad, "Nature Inspired Techniques for Data Clustering," Circuits, Systems, Communication and Information Technology Applications (CSCITA), 2014 International Conference, pp. 419 - 424, 2014.

2. X. Cui and F. Wang, "An Improved Method for K-Means Clustering," *2015 International Conference on Computational Intelligence and Communication Networks (CICN),* pp. 756 - 759, 2015.

3. K. Mizuno, S. Takamatsu, T. Shimoyama and S. Nishihara, "Fireflies can find groups for data clustering," *2016 IEEE International Conference on Industrial Technology (ICIT),* pp. 746 - 751, 2016.

4. M. R. Meybodi and T. Hassanzadeh, "A new hybrid approach for data clustering using firefly algorithm and K-means," *Artificial Intelligence and Signal Processing (AISP), 2012 16th CSI International Symposium,* pp. 007 - 011, 2012.

5. L. Parthiban and G. George, "Multi objective hybridized firefly algorithm with group search optimization for data clustering," *2015 IEEE International Conference on Research in Computational Intelligence and Communication Networks (ICRCICN),* pp. 125 - 130, 2015.

6. M. R. Meybodi and T. Hassanzadeh, "A new hybrid approach for data clustering using firefly algorithm and K-means," *Artificial Intelligence and Signal Processing (AISP), 2012 16th CSI International Symposium,* pp. 7 - 11, 2012.

7. M. Çelik, D. Karaboga and F. Koylu, "Artificial bee colony data miner (ABC-Miner)," *Innovations in Intelligent Systems and Applications (INISTA), 2011 International Symposium,* pp. 96 - 100, 2011.

8. D. Karaboga, C. Ozturk and E. Hancer, "Artificial Bee Colony based image clustering method," *2012 IEEE Congress on Evolutionary Computation,* pp. 1 - 5, 2012.

9. E. Mezura-Montes and O. Cetina-Domínguez, "Empirical Analysis of a Modified Artificial Bee Colony for Constrained Numerical Optimization," 2012.

10. M. B. Bonab and S. Z. Mohd Hashim, "Improved k-means clustering with Harmonic-Bee algorithms," *Information and Communication Technologies (WICT), 2014 Fourth World Congress,* pp. 332 - 337, 2014

11. D. Liyan, Z. Sainan, T. Geng, L. Yongli and C. Guanyan, "Ant Colony Clustering Algorithm Based on Swarm Intelligence," *International Conference on Intelligent Networks and Intelligent Systems,* pp. 123 - 126, 2013.

12. Gnanapriya and P. S. Ranjani, "Initialization K-Means Using Ant Colony Optimization," *International Journal of Engineering Research and Science & Technology,* vol. 2, pp. 85 - 92, 2013.

13. H. Fu, "A Novel Clustering Algorithm With Ant Colony Optimization," *Computational Intelligence and Industrial Application, 2008. PACIIA '08. Pacific-Asia Workshop,* vol. 2, pp. 66 - 69, 2008.

14. M. K. Nair and K. Aparna, "Enhancement of K-Means algorithm using ACO as an optimization technique on high dimensional data," *Electronics and Communication Systems (ICECS), 2014 International Conference,* pp. 1 - 5, 2014.

15. S. Mehta and P. Agarwal, "Comparative analysis of nature inspired algorithms on data clustering," *2015 IEEE International Conference on Research in Computational Intelligence and Communication Networks (ICRCICN),* pp. 119 - 124, 2015.

16. C. Deng, Z. Wang, Z. Wu and D. C. Tran, "A Novel Hybrid Data Clustering Algorithm Based on Artificial Bee Colony Algorithm and K-Means," *Chinese Journal of Electronics,* vol. 24, no. 4, pp. 694 - 701, 2015.

17. D. Maji, M. Biswas, G. Sarkar and I. Dey, "Hierarchical Clustering for Segmenting Fused Image Using Discrete Cosine Transform with Artificial Bee Colony Optimization," *2016 Second International Conference on Computational Intelligence & Communication Technology (CICT),* pp. 54 - 59, 2016.

18. V. Snasel, J. Platos and J. Janousek, "Clustering using artificial bee colony on CUDA," *2014 IEEE International Conference on Systems, Man, and Cybernetics (SMC),* pp. 3803 - 3807, 2014.

19. C. I. Mary and S. K. Dr. Raja, "Refinement Of Clusters From K-Means With Ant Colony Optimization," *Journal of Theoretical and Applied Information Technology,* pp. 28 - 32, 2005.

20. D. A. B. Pereira, A. C. B. K. Vendramin, A. T. R. Pozo, W. K. G. Assuncao and T. E. Colanzi, "Empirical Studies on Application of Genetic Algorithms and Ant Colony Optimization for Data Clustering," *Chilean Computer Science Society (SCCC), 2010 XXIX International Conference,* pp. 1 - 10, 2010.

21. S. Xu, Z. Bing, Y. Lina, L. Shanshanl and G. Lianru, "Hyperspectal Image Clustering Using Ant Colony Optimization(ACO) Improved by K-means Algorithm," *International Conference on Advanced Computer Theory and Engineering(ICACTE),* vol. 2, pp. 474 - 478, 2010.

22. P. R. P. Naga and U. Chandrasekhar, "Recent Trends in Ant Colony Optimization and Data Clustering: A Brief Survey," *Intelligent Agent and Multi-Agent Systems (IAMA), 2011 2nd International Conference,* pp. 32 - 36, 2011.

23. S.-C. Chu, J. F. Roddick, Che-Jen Su and J.-S. Pan, "Constrained Ant Colony Optimization for Data Clustering," pp. 534 - 543.

24. R. Jensi and G. W. Jiji, "A New Data Clustering Method Using Modified Bat Algorithm And Levy Flight," *ICTACT Journal ON Soft Computing,* vol. 06, no. 01, pp. 1093 - 1101, 2015.

25. P. Neelavathi, K. Thangavel and E. S. Kumar, "Performance Analysis of BAT K-Means Clustering Algorithm Using Gene Expression Data Set," *International Journal of Computational Intelligence and Informatics,* vol. 5, pp. 359 - 363, 2016.